\documentclass{article}
\usepackage{graphicx} 
\usepackage[final]{neurips_2025_ml4ps}
\usepackage{amsmath}
\usepackage{amssymb}
\usepackage[hidelinks]{hyperref}
\usepackage{tabularx}
\usepackage[table]{xcolor}
\newcolumntype{Y}{>{\raggedleft\arraybackslash}X} 
\title{Direct Molecular Polarizability Prediction with $\boldsymbol{SO(3)}$ Equivariant Local Frame GNNs}

\author{
  Jean Philip Filling\textsuperscript{1} \quad
  Felix Post\textsuperscript{2} \quad
  Michael Wand\textsuperscript{1} \quad
  Denis Andrienko\textsuperscript{2}
  \\
  \textsuperscript{1}Institute of Computer Science, Johannes Gutenberg-University, Mainz \\
  \textsuperscript{2}Max Planck Institute for Polymer Research, Mainz \\
  \\
  jefillin@uni-mainz.de \quad postf@mpip-mainz.mpg.de
}

\begin{document}

\maketitle

\begin{abstract}
We introduce a novel equivariant graph neural network (GNN) architecture designed to predict the tensorial response properties of molecules. Unlike traditional frameworks that focus on regressing scalar quantities and derive tensorial properties from their derivatives, our approach maintains $SO(3)$-equivariance through the use of local coordinate frames. Our GNN effectively captures geometric information by integrating scalar, vector, and tensor channels within a local message-passing framework. To assess the accuracy of our model, we apply it to predict the polarizabilities of molecules in the QM7-X dataset and show that tensorial message passing outperforms scalar message passing models. This work marks an advancement towards developing structured, geometry-aware neural models for molecular property prediction.
\end{abstract}

\section{Introduction}
Molecular polarizability characterizes the response of a molecule's electronic cloud to external fields and governs intermolecular interactions and dielectric behavior. Although density functional theory (DFT) can be very accurate, it is costly for large systems, motivating machine learning (ML) surrogates. For geometric molecular data, respecting symmetry is crucial: predictions should transform predictably under rotations (equivariance). This can be encouraged by data augmentation or enforced by equivariant architectures, the latter are often more data-efficient \cite{cohen2016groupequivariantconvolutionalnetworks} \cite{cohen2019gaugeequivariantconvolutionalnetworks}. We propose an $SO(3)$-equivariant message-passing GNN that uses local reference frames to directly predict $3{\times}3$ polarizability tensors, combining scalar, vector, and tensor channels to capture directional interactions.

\section{Related Work}
Early ML models such as SchNet~\cite{schutt2017schnet}, PhysNet~\cite{unke2019physnet}, and DimeNet~\cite{gasteiger2020fast} achieve strong accuracy on scalar targets (e.g., energies). Many observables of interest, however, are tensorial. Equivariant architectures—including Tensor Field Networks~\cite{thomas2018tensor}, Cormorant~\cite{anderson2019cormorant}, SE(3)-Transformers~\cite{fuchs2020se}, and NequIP~\cite{batzner20223}—use spherical harmonics and Clebsch–Gordan tensor products to encode rotation symmetry and can in principle represent high-rank quantities. In practice, tensors are often obtained as derivatives of learned scalars. In contrast, we aim to directly regress rank-2 tensors. We adopt a node-local, frame-aware message passing formalism in which each atom, represented as a node, carries its own local frame for representing and exchanging tensorial features, following the idea introduced in~\cite{lippmann2025beyond}. The key difference is that in \cite{lippmann2025beyond}, all tensorial quantities are updated jointly through a single MLP,
whereas our model maintains separate scalar, vector, and tensor channels and allows explicit control over their interactions,
for example through tensor products or gated mixing.

\section{Method}

\paragraph{Local Reference Frames.}
A local reference frame (LF) assigns each atom (node) in a molecular graph to an orthonormal basis constructed from its neighborhood. Expressing directional or tensorial quantities in these frames preserves the desired symmetries: the construction is translation invariant and $SO(3)$-equivariant with respect to rotations, enabling GNNs to process geometry effectively.

Prior work proposed both learned and classical frames. Wang et al.~\cite{wang2022graph} learn weights for neighbor directions within a cutoff and obtain an orthonormal basis via Gram--Schmidt. Lippmann et al.~\cite{lippmann2025beyond} show that such learned frames preserve GNN expressiveness and that simple principal component analysis (PCA) can perform comparably.

\paragraph{Charge-weighted PCA frame.}
In a nuclear charge-weighted variant we set $w_{ij}=|Z_j|$ for each neighbor $j$ of node $i$ and define the weighted mean direction
\[
\boldsymbol\mu_i=\frac{\sum_j w_{ij}(\mathbf r_j-\mathbf r_i)}{\sum_j w_{ij}}
=\frac{\sum_j w_{ij}\mathbf d_{ij}}{\sum_j w_{ij}}.
\]

Here $\mathbf r_i$ denotes the vector to the ith atom, $\mathbf d_{ij} = \mathbf r_j - \mathbf r_i$ is the relative vector from atom $i$ to $j$. 
The weighted covariance is
\[
\mathbf C_i=\frac{\sum_j w_{ij}\,\mathbf d_{ij}\mathbf d_{ij}^{\top}}{\sum_j w_{ij}}-\boldsymbol\mu_i\boldsymbol\mu_i^{\top}.
\]

Let $\mathbf C_i=\mathbf E_i\boldsymbol\Lambda_i\mathbf E_i^{\top}$ be its eigendecomposition with ascending eigenvalues. The raw $z$-axis is the eigenvector of the smallest eigenvalue, $\tilde{\mathbf z}_i=\mathbf E_i(:,1)$. We obtain a deterministic orientation by aligning it with the local charge-weighted mean,
\[
\mathbf z_i=\operatorname{sign}\!\bigl(\tilde{\mathbf z}_i^{\top}\boldsymbol\mu_i\bigr)\,\tilde{\mathbf z}_i.
\]
This cue depends only on atoms within the cutoff, is translation-invariant,
and transforms as a vector, thus preserving $SO(3)$-equivariance.
If $\|\boldsymbol\mu_i\|\le\varepsilon$ or the sign test is unstable,
the principal eigenvector $\mathbf E_i(:,3)$ (largest variance direction)
is used as a fallback for the $x$-axis.
Here, $\varepsilon$ defines the threshold for switching to the fallback,
and $\boldsymbol\mu_i=0$ in perfectly symmetric neighborhoods.

For the $x$-axis we use $\boldsymbol\mu_i$ if $\|\boldsymbol\mu_i\|>\varepsilon$; otherwise we take $\mathbf E_i(:,3)$. We project to the plane orthogonal to $\mathbf z_i$ and normalize,
\[
\mathbf x_i=\frac{\boldsymbol\mu_i-(\boldsymbol\mu_i\!\cdot\!\mathbf z_i)\mathbf z_i}{\bigl\|\boldsymbol\mu_i-(\boldsymbol\mu_i\!\cdot\!\mathbf z_i)\mathbf z_i\bigr\|},
\qquad
\mathbf y_i=\frac{\mathbf z_i\times\mathbf x_i}{\|\mathbf z_i\times\mathbf x_i\|}.
\]
Finally, we re-orthonormalize and assemble
\[
\mathbf F_i=(\,\mathbf x_i\;\mathbf y_i\;\mathbf z_i\,)\in SO(3).
\]

Since the cross product enforces a right-handed basis, only proper rotations are represented, not reflections (see O(3)-equivariant constructions~\cite{lippmann2025beyond}). For highly symmetric, linear, or planar graphs the computation can degenerate and break equivariance, though this was rare in our experiments (Sec.~\ref{par:Equivariance}). Developing a robust solution remains important future work.

\paragraph{Architecture}
\begin{figure}[!htbp]
  \centering
  \vspace{-0.5em} 
  \includegraphics[width=\linewidth]{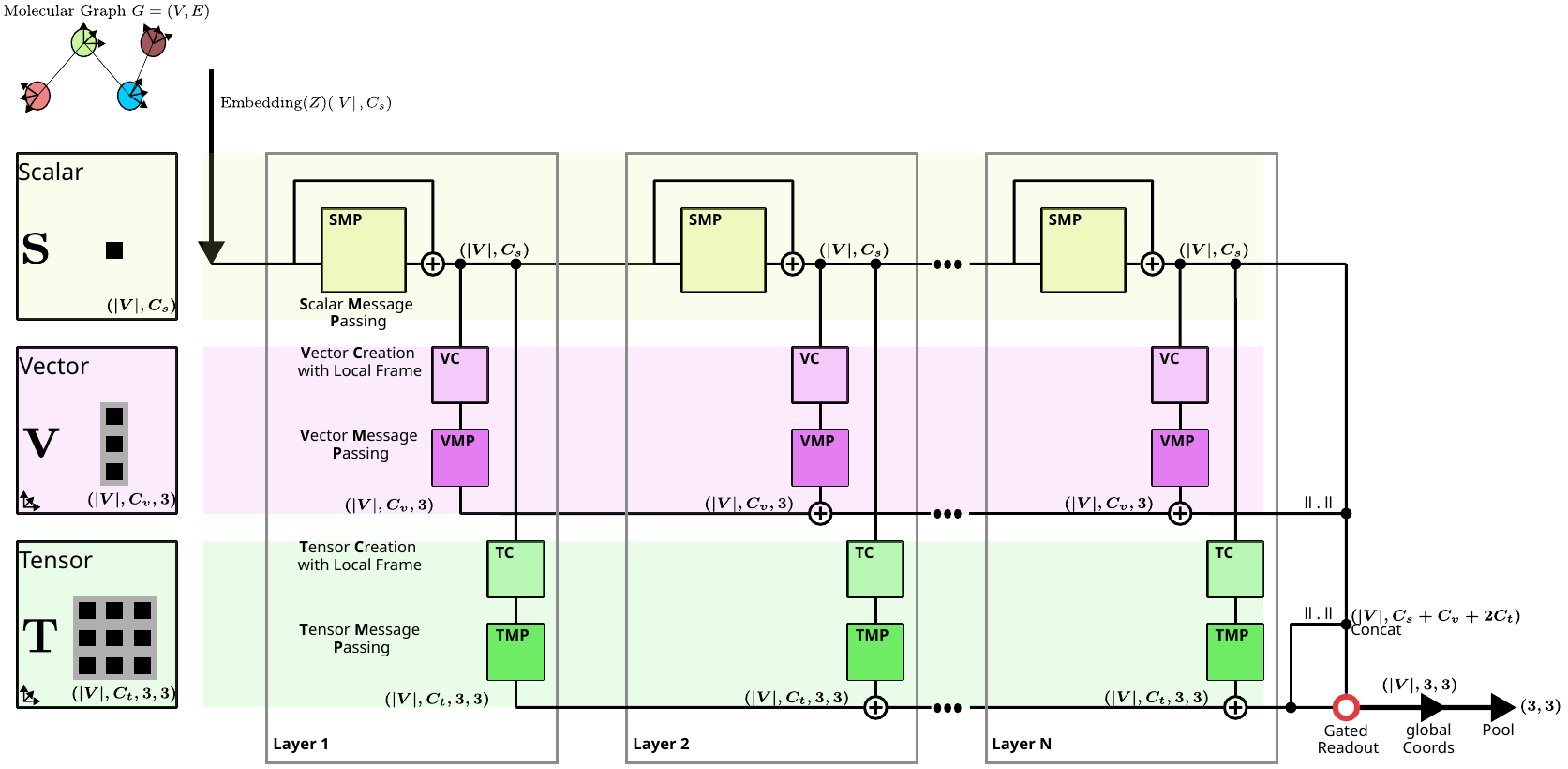}
  \vspace{-0.5em} 
  \caption{\textbf{Equivariant Architecture for Rank-2 Tensors}}
  \vspace{-0.75em} 
  \label{fig:architecture}
\end{figure}
We operate on molecular graphs with cutoff $r_c$ and attach a local frame $\mathbf{F}_i\in \mathbb{R}^{3 \times 3}$ to each atom $i$. Atomic numbers are embedded as $\mathbf{s}_i^{0}(Z_i)\in\mathbb{R}^{C_s}$;  node features are scalars $\mathbf{s}_i\in\mathbb{R}^{C_s}$, vectors $\mathbf{v}_i\in\mathbb{R}^{C_v\times 3}$, and rank-2 tensors $\mathbf{T}_i\in\mathbb{R}^{C_t\times 3\times 3}$, where $C_s, C_v \,\text{and} \, C_t$ denote the dimensions of the channels. Features are transported between nodes via the relative rotation $\mathbf{F}_{ij}=\mathbf{F}_i^{\top}\mathbf{F}_j\in SO(3)$, which preserves $SO(3)$-equivariance. The scalar backbone is an invariant MPNN that uses only rotation-invariant edge inputs $\mathbf{r}_{ij}=\mathbf{x}_i-\mathbf{x}_j$, $d_{ij}=\|\mathbf{r}_{ij}\|$, $d_{ij}^{2}$, and $(d_{ij}^{2}+\varepsilon)^{-1}$ with sigmoid-gated messages and residual updates of $\mathbf{s}_i$. Vectors and tensors are initialized from $\mathbf{s}_i$ in the local frame and passed along edges by channel-wise transport and mixing:
\[
\mathbf{v}_j^{(i)}=\mathbf{F}_{ij}\,\mathbf{v}_j,\qquad
\mathbf{T}_j^{(i)}=\mathbf{F}_{ij}\,\mathbf{T}_j\,\mathbf{F}_{ij}^{\top},
\]
followed by a per-edge interaction MLP (sigmoid output), neighborhood aggregation, and residual updates; shapes are preserved ($\mathbf{v}_i\!\in\!\mathbb{R}^{C_v\times 3}$, $\mathbf{T}_i\!\in\!\mathbb{R}^{C_t\times 3\times 3}$). The readout predicts a local $3{\times}3$ contribution per node and pools in global coordinates,
\[
\hat{\boldsymbol{\alpha}}=\mathrm{pool}_i\,\mathbf{F}_i\,\hat{\boldsymbol{\alpha}}^{\mathrm{loc}}_i\,\mathbf{F}_i^{\top},
\]
yielding an $SO(3)$-equivariant molecular polarizability by construction with gated nonlinearities adapted from \cite{weiler20183d}. For details, have a look at Sec. \ref{app:layer design}.

\section{Experiment}

\paragraph{Dataset}

We use the \textbf{QM7-X} \cite{hoja2021qm7} dataset, containing $\sim$4.2M optimized conformations of $\sim$6{,}900 molecules computed at the DFT-PBE0+MBD level. We select only optimized structures (\texttt{*-opt}), remove duplicates by molecule ID, and split the data into \textbf{80/10/10\%} train/validation/test with a fixed seed, ensuring all conformations of a molecule remain in the same split. As no baselines exist for molecular polarizabilities on QM7-X, we report only consistent intra-dataset comparisons.

\paragraph{Setup}
We compare two $SO(3)$-equivariant architectures. 
Baseline (Scalar+LF head): message passing uses only a scalar channel. After the last layer, a small tensor head decodes \emph{one} $3{\times}3$ polarizability tensor in the local frame, $\hat{\boldsymbol{\alpha}}^{\mathrm{loc}}_i$ (one tensor per atom). No tensor features or messages are exchanged across edges. The molecular prediction is obtained by rotating to global coordinates and pooling,
$\hat{\boldsymbol{\alpha}}=\mathrm{pool}_i\,\mathbf{F}_i\,\hat{\boldsymbol{\alpha}}^{\mathrm{loc}}_i\,\mathbf{F}_i^\top$.
Tensorial (S/V/T): in addition to scalars, the network maintains vector and rank-2 tensor channels. On each edge $(i,j)$, features are transported between local frames via the relative rotation $\mathbf{F}_{ij}=\mathbf{F}_i^\top\mathbf{F}_j$, and per-edge interaction multi-layer perceptrons (MLPs) mix tensorial messages before aggregation, see Fig. \ref{fig:architecture}. The same equivariant readout as above yields the molecular tensor. This isolates the effect of tensorial message passing versus a scalar-only backbone with a per-node tensor head.

\paragraph{Implementation Details}
Both models are trained for 1{,}000 epochs with a batch size of 32 and a learning rate of \(10^{-4}\). To ensure comparability, we choose layer widths such that the total number of trainable parameters is similar. The scalar model uses 331 scalar channels (\(\approx 5{,}471{,}127\) parameters), whereas the tensorial model uses 128 scalar, 4 vector, and 32 tensor channels (\(\approx 5{,}477{,}145\) parameters).We keep the number features constant over the 8 used layers. Each MLP has two layers, and the interaction layers use a sigmoid activation to implement gating. The cutoff radius $r_c$ is set to 4\,\textup{\AA}. We trained a model for each metric. See the code at \href{https://github.com/phil-fill/DiMolPol}{github.com/phil-fill/DiMolPol}.

\par\noindent
\begin{table}[ht]
\centering
\small
\begin{tabularx}{\linewidth}{|l|Y|Y|Y|}
\hline
\textbf{Metric} & \textbf{Ground Truth Scales (Mean)} & \textbf{Scalar Model + LF Head MAE} & \textbf{Tensorial Message Model MAE} \\
\hline
Tensor     &  $36.14$ & $0.53$ & $\mathbf{0.45}$ \\
Trace      & $266.61$ & $0.41$ & $\mathbf{0.36}$ \\
Anisotropy &  $11.01$ & $0.52$ & $\mathbf{0.42}$ \\
Frobenius  & $159.75$ & $1.78$ & $\mathbf{1.67}$ \\
\hline
\end{tabularx}
\caption{MAEs for different metrics of the polarizability tensor $\hat{\boldsymbol{\alpha}}$ averaged over molecules in $[bohr^{3}]$. We show the ground truth scales and results for the Scalar Model with local frame Head and the Tensorial Message Model.}
\label{tab:scales_mae}
\end{table}
\paragraph{Results}
We report four test-set metrics for both models: Tensor MAE (Mean Absolute Error) over tensor components, Trace MAE, Frobenius MAE, and Anisotropy MAE (off-diagonal entries) averaged over molecules. For all four metrics the tensorial model with message passing was superior to the simple scalar model. As both models are equivariant by construction, these differences are due to the specifics of the architecture chosen, i.e. our tensorial message passing scheme.

\paragraph{Equivariance Tests}\label{par:Equivariance}
Both models are by construction $SO(3)$-equivariant. We test this with two protocols: (i) \textbf{Model Equivariance}, where inputs (positions, frames) are rotated consistently and predictions compared to $\boldsymbol{R}\,\hat{\boldsymbol{\alpha}}^{(\text{base})}_{\text{pred}}\boldsymbol{R}^\top$; and (ii) \textbf{Pipeline Equivariance}, where positions are rotated but frames are recomputed, capturing end-to-end robustness including PCA sign flips or degeneracies. The relative Frobenius error, averaged over molecules and random rotations, serves as evaluation metric.

\[
\mathrm{rel\_frob} =
\frac{
\bigl\lVert 
\hat{\boldsymbol{\alpha}}^{(\boldsymbol{R})}_{\text{pred}}
-
\mathbf{R}\,\hat{\boldsymbol{\alpha}}^{(\text{base})}_{\text{pred}}\,\mathbf{\boldsymbol{R}}^{\top}
\bigr\rVert_F
}{
\tfrac12\!\Bigl(
\bigl\lVert \hat{\boldsymbol{\alpha}}^{(\boldsymbol{R})}_{\text{pred}} \bigr\rVert_F
+
\bigl\lVert \hat{\boldsymbol{\alpha}}^{(\text{base})}_{\text{pred}} \bigr\rVert_F
\Bigr)
+\varepsilon
}
\]

\begin{table}[ht]
\centering
\small
\begin{tabular}{|l|c|c|}
\hline
\textbf{Model} & \textbf{Model Equivariance} & \textbf{Pipeline Equivariance} \\
\hline
Scalar Model + LF Head & $(1.13 \pm 0.04)\times10^{-7}$ & $(2.93 \pm 1.70)\times10^{-5}$ \\
Tensorial Message Model & $(1.29 \pm 0.15)\times10^{-7}$ & $(2.24 \pm 0.36)\times10^{-5}$ \\
\hline
\end{tabular}
\caption{Mean and standard deviation of relative Frobenius norm as equivariance measure (64 random rotations, averaged over molecules)}
\label{tab:eq_rel_frob}
\end{table}
Model equivariance holds up to numerical noise, but end-to-end equivariance is only approximate due to local frame recomputation. The relative Frobenius error rises from $\sim 10^{-7}$ (model-only) to $\sim 10^{-5}$ (pipeline), which is still small for our dataset. However, in graphs with many nearly collinear or coplanar neighborhoods, PCA-based frames can flip or become ambiguous, amplifying the deviation. Thus, the architecture itself is $SO(3)$-equivariant, while frame recomputation is the main source of residual error.

\section{Discussion and Outlook}
In this work we introduced an $SO(3)$-equivariant GNN and demonstrated that tensorial message passing consistently improves accuracy across all reported metrics compared to a scalar baseline. Both models are by construction $SO(3)$-equivariant, and our equivariance tests confirm that deviations remain limited to numerical noise. The residual errors we observe are due to PCA-based frame ambiguities in (near-)linear or planar neighborhoods rather than shortcomings of the underlying architecture. While these effects are relatively rare, they highlight the importance of developing more robust frame definitions. 

The improvements hold consistently across all reported metrics. As future work, we plan to address the remaining limitations of the PCA-based frame construction by developing more robust variants and exploring whether frames can be learned jointly with the model. In addition, we will conduct systematic ablation studies to disentangle the contribution of scalar, vector, and tensor channels. A broader benchmarking against other state-of-the-art equivariant GNNs is also planned, in order to better position our approach within the landscape of equivariant architectures. Finally, we aim to extend our evaluation to larger and chemically more diverse datasets, which will allow us to assess both the generalization ability and the practical applicability of the proposed method.

\paragraph{Acknowledgement}
This project was funded by the Deutsche Forschungsgemeinschaft (DFG, German Research Foundation) in the framework to the collaborative research center   Multiscale Simulation Methods for Soft-Matter Systems" (TRR 146) under Project No. 233630050. We also like to thank the open source community for projects like Pytorch Geometric \cite{fey2019fast}.


\bibliographystyle{unsrt}  
\bibliography{references} 

\section{Appendix}
\subsection{Layer Design}\label{app:layer design}

\paragraph{Notation and shapes}
We consider molecular graphs $G=(V,E)$ with cutoff $r_c$.
Atoms $i\in V$ have positions $\mathbf{x}_i\in\mathbb{R}^3$ and atomic numbers $Z_i$.
Each layer $\ell=0,\dots,L-1$ maintains per node
\[
\mathbf{s}_i^{(\ell)}\in\mathbb{R}^{C_s}\quad(\text{scalars}),\qquad
\mathbf{v}_i^{(\ell)}\in\mathbb{R}^{C_v\times 3}\quad(\text{vectors}),\qquad
\mathbf{T}_i^{(\ell)}\in\mathbb{R}^{C_t\times 3\times 3}\quad(\text{rank-2 tensors}).
\]
Each node carries a local orthonormal frame $\mathbf{F}_i\in SO(3)$. Relative rotations are
\[
\mathbf{F}_{ij}=\mathbf{F}_i^{\top}\mathbf{F}_j\in SO(3).
\]
Channel-wise concatenation is denoted as$ [\cdot,\cdot]$. All MLPs are two-layer. Interaction MLPs use a sigmoid on the last layer to output coefficients in $(0,1)$.

\paragraph{Initialization}
Atomic numbers are embedded to scalars $\mathbf{s}_i^{0} \in\mathbb{R}^{C_s}$ from a lookup table of $Z_i$ {are taken from \ldots (e.g. NIST)}.

\subsection*{Scalar channel ($S$)}
The backbone is a usual message passing GNN where the interaction is calculated over the nearest neighbors specified with the cutoff radius $r_c$. We model the scalar channel in a complete invariant way and feed only invariants into the different MLPs.
For $(i,j)\in E$ let $\mathbf{r}_{ij}=\mathbf{x}_i-\mathbf{x}_j$, $d_{ij}^2=\|\mathbf{r}_{ij}\|^2$, $d_{ij}^{-2}=(d_{ij}^2+\varepsilon)^{-1}$.
\[
\mathbf{m}_{ij}^{(\ell)}=\phi_e\bigl([\mathbf{s}_i^{(\ell)},\mathbf{s}_j^{(\ell)},d_{ij}^2,d_{ij}^{-2}]\bigr)\in\mathbb{R}^{C_s},\quad
g_{ij}^{(\ell)}=\sigma\!\bigl(\phi_{g}(\mathbf{m}_{ij}^{(\ell)})\bigr)\in(0,1)
\]
From the messages we apply another MLP with a sigmoid function to gate the information flow of the actual message. Afterwards the message will be aggregated at node $i$ over nearest neighbors and then updated. The MLPs are constructed in a way that the depth of the network does not change over layers.
\[
\mathbf{m}_i^{(\ell)}=\sum_{j\in\mathcal{N}(i)} g_{ij}^{(\ell)}\cdot\mathbf{m}_{ij}^{(\ell)},\qquad
\mathbf{s}_i^{(\ell+1)}=\mathbf{s}_i^{(\ell)}+\phi_s\bigl([\mathbf{s}_i^{(\ell)},\mathbf{m}_i^{(\ell)}]\bigr)
\]
For the edge inferring we used a sigmoid to gate the different channels, like the EGNN paper \cite{satorras2021n}. After the first update the scalar channel is used to create higher tensorial properties in the local frame of each node.

\[
\mathbf{v}_i^{(0)}=\mathrm{reshape}\bigl(\phi_v^{\mathrm{init}}(\mathbf{s}_i^{(1)})\bigr)\in\mathbb{R}^{C_v\times 3},\qquad
\quad
\mathbf{T}_i^{(0)}=\mathrm{reshape}\bigl(\phi_t^{\mathrm{init}}(\mathbf{s}_i^{(1)})\bigr)\in\mathbb{R}^{C_t\times 3\times 3}
\]
\subsection*{Vector channel ($V$)}

\textbf{Vector Initialization:} For the Vector Initialization a MLP is trained to take the scalar channel and produce vectors in the local frames. 
\[
\tilde{\mathbf{v}}_{i}^{(\ell + 1)}=\mathrm{reshape}\bigl(\phi_v(\mathbf{s}_i^{(\ell)})\bigr)\in\mathbb{R}^{C_v\times 3}
\]

\textbf{Vectorial Message Passing:} For each edge $(i,j)$ and channel the sender vectors are rotated into the receiver frame:
\[
\tilde{\mathbf{v}}_j^{(i,\ell + 1)}[c]=\mathbf{F}_{ij}\,\tilde{\mathbf{v}}_j^{(\ell + 1)}[c]\quad \forall c\in\{1,\dots,C_v\}
\]
We are stacking vectors from neighbors $i,j$ represented in the local frame of $i$. We first train a matrix that learns a gate weight for each channel $U_{ij}^{(\ell)}=[\,\tilde{\mathbf{v}}_i^{(\ell + 1)},\tilde{\mathbf{v}}_j^{(i,\ell + 1)}\,]\in\mathbb{R}^{(2C_v)\times 3}$.
\[
\mathbf{W}_{ij}^{(v,\ell)}=\sigma\!\bigl(\phi_{v,\mathrm{int}}([\mathbf{s}_i^{(\ell)},\mathbf{s}_j^{(\ell)}])\bigr)\in(0,1)^{C_v\times 2C_v}
\]
Now the interaction takes place. For every new channel there are $C_v$ vector scalar (gate) multiplications to form a new channel as a superposition from the old ones. 
\[
\mathbf{v}_{ij}^{\mathrm{mix},(\ell)}[c,:]=\sum_{k=1}^{2C_v}\mathbf{W}_{ij}^{(v,\ell)}[c,k]\;U_{ij}^{(\ell)}[k,:]\in\mathbb{R}^{1\times 3}
\]
We can update the channel using
\[
\mathbf{v}_i^{(\ell+1)}=\mathbf{v}_i^{(\ell)}+\sum_{j\in\mathcal{N}(i)}\mathbf{v}_{ij}^{\mathrm{mix},(\ell)}\in\mathbb{R}^{C_v\times 3}.
\]

\subsection*{Tensor channel ($T$, rank-2)}

\textbf{Tensor Initialization:} Analogous to vectors, a small MLP takes the scalar state and produces rank-2 tensors (in local frames):
\[
\tilde{\mathbf{T}}_{i}^{(\ell+1)}=\mathrm{reshape}\bigl(\phi_t(\mathbf{s}_i^{(\ell)})\bigr)\in\mathbb{R}^{C_t\times 3\times 3}.
\]

\textbf{Tensorial Message Passing:} For each edge $(i,j)$, rotate sender tensors into the receiver frame per channel:
\[
\tilde{\mathbf{T}}_{j}^{(i,\ell+1)}[c]=\mathbf{F}_{ij}\,\tilde{\mathbf{T}}_{j}^{(\ell+1)}[c]\,\mathbf{F}_{ij}^{\top}
\quad \forall c\in\{1,\dots,C_t\}.
\]
We stack tensors from $i$ and $j$ expressed in $i$’s frame,
\[
U_{ij}^{(\ell)}=\bigl[\,\tilde{\mathbf{T}}_{i}^{(\ell+1)},\;\tilde{\mathbf{T}}_{j}^{(i,\ell+1)}\,\bigr]\in\mathbb{R}^{(2C_t)\times 3\times 3}.
\]
A per-edge interaction MLP with sigmoid output produces channel mixing coefficients
\[
\mathbf{W}_{ij}^{(t,\ell)}=\sigma\!\bigl(\phi_{t,\mathrm{int}}([\mathbf{s}_i^{(\ell)},\mathbf{s}_j^{(\ell)}])\bigr)\in(0,1)^{C_t\times 2C_t}.
\]
The interaction forms new channels as superpositions of the stacked inputs (broadcast over $3\times 3$):
\[
\mathbf{T}_{ij}^{\mathrm{mix},(\ell)}[c,:,:]=\sum_{k=1}^{2C_t}\mathbf{W}_{ij}^{(t,\ell)}[c,k]\;U_{ij}^{(\ell)}[k,:,:]
\]
\textbf{Aggregation and update} Finally, aggregate over neighbors and apply a residual update:
\[
\mathbf{T}_i^{(\ell+1)}=\mathbf{T}_i^{(\ell)}+\sum_{j\in\mathcal{N}(i)}\mathbf{T}_{ij}^{\mathrm{mix},(\ell)}
\;\in\;\mathbb{R}^{C_t\times 3\times 3}
\]

\subsection*{Equivariance sketch}
Under a global rotation $\mathbf{R}\in SO(3)$, positions transform as $\mathbf{x}_i\mapsto \mathbf{R}\mathbf{x}_i$.
The frame construction implies $\mathbf{F}_i\mapsto \mathbf{R}\mathbf{F}_i$ (translation invariant, $SO(3)$-equivariant), hence
\[
\mathbf{F}_{ij}=\mathbf{F}_i^{\top}\mathbf{F}_j\;\mapsto\;(\mathbf{R}\mathbf{F}_i)^{\top}(\mathbf{R}\mathbf{F}_j)=\mathbf{F}_{ij},
\]
so all transports are rotation-consistent.
Local predictions transform as $\hat{\boldsymbol{\alpha}}_i^{\mathrm{loc}}\mapsto \hat{\boldsymbol{\alpha}}_i^{\mathrm{loc}}$ (computed in local coordinates),
and the readout gives
\[
\hat{\boldsymbol{\alpha}}\mapsto \mathrm{pool}_i\;(\mathbf{R}\mathbf{F}_i)\hat{\boldsymbol{\alpha}}_i^{\mathrm{loc}}(\mathbf{R}\mathbf{F}_i)^{\top}
=\mathbf{R}\bigl(\mathrm{pool}_i\;\mathbf{F}_i\hat{\boldsymbol{\alpha}}_i^{\mathrm{loc}}\mathbf{F}_i^{\top}\bigr)\mathbf{R}^{\top}
=\mathbf{R}\,\hat{\boldsymbol{\alpha}}\,\mathbf{R}^{\top}.
\]

\end{document}